\documentclass[a4paper,11pt,twocolumn,twoside]{article}
\usepackage[dvips]{graphicx}
\usepackage{sepln_en}
\usepackage{fullname}
\usepackage[utf8]{inputenc}
\usepackage{tabularx} 
\usepackage{graphicx}
\usepackage{booktabs}
\usepackage[english]{babel}
\usepackage{url}

\input epsf

\title{Overview of ADoBo 2021: Automatic Detection of Unassimilated Borrowings in the Spanish Press}

\author {\textbf{Elena Álvarez Mellado$^1$,} \textbf{Luis Espinosa Anke$^2$,} \textbf{Julio Gonzalo Arroyo$^1$,} \\ \textbf{Constantine Lignos$^3$,} \textbf{Jordi Porta Zamorano$^4$}\\
$^1$NLP \& IR group, UNED, Madrid, Spain\\
$^2$School of Computer Science and Informatics, Cardiff University, Cardiff, UK\\
$^3$Michtom School of Computer Science, Brandeis University, Massachusetts, USA\\
$^4$Centro de Estudios de la RAE, Madrid, Spain\\

elena.alvarez@lsi.uned.es, espinosa-ankel@cardiff.ac.uk,\\ julio@lsi.uned.es, lignos@brandeis.edu, porta@rae.es\\
}

\seplntranstitle{Resumen de ADoBo 2021: detección automática de préstamos léxicos no asimilados en la prensa española}

\seplnclave{Préstamo léxico, anglicismos, detección automática de préstamos.}

\seplnresumen{En este artículo presentamos los resultados de ADoBo 2021, la tarea compartida de IberLEF 2021 sobre detección de préstamos léxicos en la prensa española. En esta tarea abordamos la detección de préstamos como un problema de etiquetado de secuencias. A los participantes de la tarea se les proporcionó un corpus de prensa española anotado con préstamos léxicos no asimilados (mayoritariamente anglicismos) siguiendo el esquema BIO. Recibimos nueve sistemas distintos provenientes de cuatro equipos diferentes. Los resultados obtenidos oscilan entre los 37 y los 85 puntos de valor F1, lo que indica que la detección de préstamos léxicos es un problema no resuelto (sobre todo cuando se abordan préstamos no vistos anteriormente) y que el trabajo lexicográfico tradicional podría beneficiarse de incorporar las técnicas actuales del PLN.}

\seplnkey{Automatic detection of borrowings, loanword detection, linguistic borrowing, anglicisms.}

\seplnabstract{
This paper summarizes the main findings of the ADoBo 2021 shared task, proposed in the context of IberLef 2021. In this task, we invited participants to detect lexical borrowings (coming mostly from English) in Spanish newswire texts. This task was framed as a sequence classification problem using BIO encoding. We provided participants with an annotated corpus of lexical borrowings which we split into training, development and test splits. We received submissions from 4 teams with 9 different system runs overall. The results, which range from F1 scores of 37 to 85, suggest that this is a challenging task, especially when out-of-domain or OOV words are considered, and that traditional methods informed with lexicographic information would benefit from taking advantage of current NLP trends.}

\firstpageno{1}

\begin{document}

% la siguiente instrucción sólo se debe usar si el abstract sobrescribe el texto
% la longitud variará según se necesite

\setlength\titlebox{20cm} % se aumenta el tamaño del espacio reservado para datos de título

\label{firstpage} \maketitle

\section{Introduction}

Lexical borrowing is the process of importing words from one language into another \cite{onysko2007anglicisms,poplack1988social}, a phenomenon that occurs in all languages. The task of automatically extracting lexical borrowings from text has proven to be relevant in lexicographic work as well as for NLP downstream tasks, such as parsing \cite{alex2008automatic}, text-to-speech synthesis \cite{leidig2014automatic} and machine translation \cite{tsvetkov2016cross}.

In recent decades, English in particular has produced numerous lexical borrowings (often called \textit{anglicisms}) in many European languages \cite{furiassi2012anglicization}. Previous work estimated that a reader of French newspapers encounters a new lexical borrowing every 1,000 words  \cite{chesley_paula_predicting_2010}, English borrowings outnumbering all other borrowings combined \cite{chesley2010lexical}. In Chilean newspapers, lexical borrowings account for approximately 30\% of neologisms, 80\% of those corresponding to anglicisms \cite{gerding2014anglicism}. In European Spanish, it was estimated that anglicisms could account for 2\% of the vocabulary used in Spanish newspaper El País in 1991 \cite{gorlach_felix}, a number that is likely to be higher today. As a result, the usage of lexical borrowings in Spanish (and particularly anglicisms) has attracted lots of attention, both in linguistic studies and among the general public. 

For ADoBo 2021, we proposed a shared task on automatically detecting lexical  borrowings in Spanish newswire, with a special focus on unassimilated anglicisms. In this paper we describe the purpose and scope of the shared task, introduce the systems that participated in it, and share the results obtained during the competition.

\section{Related work}

Several projects have approached the task of extracting lexical borrowings in various European languages, such as German \cite{alex2008automatic,alex-2008-comparing,garley-hockenmaier-2012-beefmoves,leidig2014automatic}, Italian \cite{furiassi2007retrieval}, French \cite{alex2008automatic,chesley2010lexical}, Finnish \cite{mansikkaniemi2012unsupervised}, and Norwegian \cite{andersen2012semi,losnegaard2012data}, with a particular focus on anglicism extraction. 

Despite the interest in modeling anglicism usage, the problem of automatically extracting lexical borrowings has been seldom explored in the NLP literature for Iberian languages in general and for Spanish in particular, with only a few recent exceptions \cite{serigos2017applying,alvarez2020lazaro}. 
\begin{table*}[t]
\small
\centering
\begin{tabular}{lclrrrrrr}
\toprule
\multicolumn{1}{c}{\textbf{Team}} 
& \multicolumn{1}{c}{\textbf{System}}	
& \multicolumn{1}{c}{\textbf{Type}}
& \multicolumn{1}{c}{\textbf{Prec.}}	
& \multicolumn{1}{c}{\textbf{Rec.}}	
& \multicolumn{1}{c}{\textbf{F1}}
& \multicolumn{1}{c}{\textbf{Ref.}}	
& \multicolumn{1}{c}{\textbf{Pred.}}	
& \multicolumn{1}{c}{\textbf{Corr.}} 
\\
\midrule
& &	ALL	    & 88.81	& 81.56	& 85.03	& 1,285	& 1,180	& 1,048 \\
Marrouviere   &   (1)    &	\texttt{ENG}	    & 90.70	& 82.65	& 86.49	& 1,239	& 1,129	& 1,024 \\
 &    &	\texttt{OTHER}	& 47.06	& 52.17	& 49.48	& 46	& 51	& 24 \\
\hline
 & &	ALL	& 88.77	& 81.17	& 84.80	& 1,285	& 1,175	& 1,043 \\
Versae   &   (2)    &	\texttt{ENG}	& 90.31	& 82.73	& 86.35	& 1,239	& 1,135	& 1,025 \\
   &    &	\texttt{OTHER}	& 45.00	& 39.13	& 41.86	& 46	& 40	& 18 \\
\hline
 &  &	ALL	    & 89.40	& 66.30	& 76.14	& 1,285	& 953	& 852 \\
Marrouviere   & (3)      &	\texttt{ENG}	    & 90.98	& 67.55	& 77.54	& 1239	& 920	& 837 \\
   &    &	\texttt{OTHER}	& 45.45	& 32.61	& 37.97	& 46	& 33	& 15 \\
\hline
 &  &	ALL	&92.28	&61.40	& 73.74	&1,285	&855	&789 \\
 Marrouviere   &   (4)    &	\texttt{ENG}	&93.43	&63.12	& 75.34	&1,239	&837	&782 \\
   &    &	\texttt{OTHER}	&38.89	&15.22	&21.88	&46	&18	&7 \\
\hline
&  &	ALL	& 62.76	& 46.30	& 53.29	& 1,285	& 948	& 595 \\
Versae   &   (5)   &	\texttt{ENG}	& 62.97	& 47.62	& 54.23	& 1,239	& 937	& 590 \\
   &    &	\texttt{OTHER}	& 45.45	& 10.87	& 17.54	& 46	& 11	& 5 \\
\hline
 &  &	ALL	    & 65.15	& 37.82	& 47.86	& 1,285	& 746	& 486 \\
Mgrafu   &   (6)    &	\texttt{ENG}	    & 65.31	& 38.90	& 48.76	& 1,239	& 738	& 482 \\
   &    &	\texttt{OTHER}	& 50.0	& 8.69	& 14.81	& 46	& 8	    & 4 \\
\hline
 &  &	ALL	& 75.27	& 27.47	& 40.25	& 1,285	& 469	& 353\\
BERT4EVER   &   (7)    &	\texttt{ENG}	& 75.43	& 28.25	& 41.10	& 1,239	& 464	& 350\\
   &    &	\texttt{OTHER}	& 60.00	& 6.52	& 11.76	& 46	& 5	& 3 \\
\hline
 &  &	ALL	    & 76.29	& 25.29	& 37.99	& 1,285 &	426 &	325\\
BERT4EVER   &   (8)    &	\texttt{ENG}	    & 76.48	& 25.99	& 38.80	& 1,239 & 	421 &	322\\
   &    &	\texttt{OTHER}	& 60.00	&  6.52	& 11.76	&   46 &	  5 &	  3\\
\hline
 &  &	ALL	  & 76.44	& 24.75	& 37.39	& 1,285	& 416	& 318 \\
BERT4EVER   &   (9)    &	\texttt{ENG}	  & 76.64	& 25.42	& 38.18 & 1,239	& 411	& 315 \\
   &    &	\texttt{OTHER} & 60.00	& 6.52	& 11.76	&   46	&   5	& 3 \\
\bottomrule
\end{tabular}
\caption{Results on the test set. For each label, precision, recall and F1 score are provided, along with the reference number of borrowings, the predicted number of borrowings and the number of correct predictions. }\label{tab:test}
\end{table*}

\section{Lexical borrowing: scope of the phenomenon}
\label{section:scope}

The concept of \textit{linguistic borrowing} covers a wide range of linguistic phenomena, but is generally understood as the process of introducing words, elements or patterns of one language (the donor language) into another language (the recipient language) \cite{haugen1950analysis,weinreich1963languages}. In that sense, lexical borrowing is somewhat similar to linguistic code-switching (the process of using two languages interchangeably in the same discourse that is common among bilingual speakers), and in fact both phenomena have been sometimes described as a continuum with a fuzzy frontier between the two \cite{clyne2003dynamics}. Consequently, disagreement on what a borrowing is (and is not) exists \cite{gomez1997towards} and various classifications and typologies for characterizing borrowing usage have been proposed, both for borrowings in general \cite{thomason1992language,matras2007grammatical,haspelmath2009loanwords} and for anglicism usage in Spanish in particular \cite{pratt1980anglicismo,lorenzo1996anglicismos,gomez1997towards,gonzalez1999anglicisms,nogueroles2018comprehensive}.

\section{Task description}

For the ADoBo shared task we have focused on unassimilated lexical borrowings, words from another language that are used in Spanish without orthographic modification and that have not (yet) been integrated into the recipient language---for example, \textit{running}, \textit{smartwatch}, \textit{influencer}, \textit{holding}, \textit{look}, \textit{hype}, \textit{prime time} and \textit{lawfare}. 

\begin{table}[ht]
\centering
\footnotesize
\begin{tabular}[t]{lrrrr}
\toprule
\multicolumn{1}{c}{\textbf{Set}} & 
\multicolumn{1}{c}{\textbf{Tokens}} & 
\multicolumn{1}{c}{\texttt{ENG}} & 
\multicolumn{1}{c}{\texttt{OTHER}} & 
\multicolumn{1}{c}{\textbf{Unique}} \\
\midrule
Train & 231,126  & 1,493 & 28 & 380 \\
Dev. & 82,578 & 306 & 49 & 316 \\
Test & 58,997 & 1,239 & 46 & 987 \\
\midrule
\textbf{Total} & 372,701 & 3,038 & 123 & 1,683 \\
\bottomrule
\end{tabular}
\caption{Corpus split and counts.}
\label{tab:corpus}
\end{table}

\subsection{Motivation for the task}

The task of extracting unassimilated lexical borrowings is a more challenging undertaking than it might appear to be at first. To begin with, lexical borrowings can be either single or multitoken expressions (e.g., \textit{prime time}, \textit{tie break} or \textit{machine learning}). Second, linguistic assimilation is a diachronic process and, as a result, what constitutes an unassimilated borrowing is not clear-cut. For example, words like \textit{bar} or \textit{club} were unassimilated lexical borrowings in Spanish at some point in the past, but have been around for so long in the Spanish language that the process of phonological and morphological adaptation is now complete and they cannot be considered unassimilated borrowings anymore. On the other hand, \textit{realia} words, that is, culture-specific elements whose name entered via the language of origin decades ago (like \textit{jazz} or \textit{whisky}) cannot be considered unassimilated anymore, despite their orthography not having been adapted into Spanish conventions. 

All these subtleties make the annotation of lexical borrowings non-trivial. Consequently, in prior work on anglicism extraction from Spanish text, plain dictionary lookup produced very limited results with F1 scores of 47 \cite{serigos2017applying} and 26 \cite{alvarez2020lazaro}. In fact, whether a given expression is a borrowing or not cannot always be determined by plain dictionary lookup; after all, an expression such as \textit{social media} is an anglicism in Spanish, even when both \textit{social} and \textit{media} also happen to be Spanish words that are registered in regular dictionaries. This justifies the need for a more NLP-heavy approach to the task, which has already proven to be promising. Previous work on borrowing extraction using a CRF model with handcrafted features produced an F1 score of 86 on a corpus of Spanish headlines \cite{alvarez2020lazaro}. 

Finally, although there are some already well-established shared tasks on mixed-language settings, they have focused exclusively on code-switched data \cite{solorio-etal-2014-overview,molina-etal-2016-overview,aguilar-etal-2018-named}, which is close to borrowing but different in scope and nature (see Section \ref{section:scope}), and no specific venue exists on borrowing detection in NLP so far. To the best of our knowledge, ADoBo is the first shared task specifically devoted to linguistic borrowing.

\begin{table*}[t]
\small
\centering
\begin{tabular}{lclrrrrrr}
\toprule
\multicolumn{1}{c}{\textbf{Team}} 
& \multicolumn{1}{c}{\textbf{System}}	
& \multicolumn{1}{c}{\textbf{Type}}
& \multicolumn{1}{c}{\textbf{Prec.}}	
& \multicolumn{1}{c}{\textbf{Rec.}}	
& \multicolumn{1}{c}{\textbf{F1}}
& \multicolumn{1}{c}{\textbf{Ref.}}	
& \multicolumn{1}{c}{\textbf{Pred.}}	
& \multicolumn{1}{c}{\textbf{Corr.}} 
\\
\midrule
 &  & 	ALL	& 73.66	& 82.49	& 77.83	& 1,285	& 1,439	& 1,060 \\
Marrouviere & (1) & 	\texttt{ENG}	& 76.31	& 83.45	& 79.72	& 1,239	& 1,355	& 1,034 \\
 &  & 	\texttt{OTHER}	& 30.95	& 56.52	& 40.00	& 46	& 84	& 26 \\
\hline
&  & 	ALL	& 81.49	& 63.04	& 71.08	& 1,285	& 994	& 810 \\
Marrouviere & (4)  & 	\texttt{ENG}	& 82.70	& 64.81	& 72.67	& 1,239	& 971	& 803 \\
 &  & 	\texttt{OTHER}	& 30.43	& 15.22	& 20.29	& 46	& 23	& 7 \\
\hline
&  & 	ALL	& 72.66	& 67.63	& 70.05	& 1,285	& 1,196	& 869 \\
Marrouviere &  (3)  & 	\texttt{ENG}	& 75.49	& 68.85	& 72.01	& 1,239	& 1,130	& 853 \\
 &  & 	\texttt{OTHER}	& 24.24	& 34.78	& 28.57	& 46	& 66	& 16 \\
\hline
&  & 	ALL	& 59.57	& 82.33	& 69.13	& 1,285	& 1,776	& 1,058 \\
Versae &  (2)  & 	\texttt{ENG}	& 61.34	& 84.02	& 70.91	& 1,239	& 1,697	& 1041 \\
 &  & 	\texttt{OTHER}	& 21.52	& 36.96	& 27.20	& 46	& 79	& 17 \\
\hline
&  & 	ALL	& 42.27	& 48.48	& 45.16	& 1,285	& 1,474	& 623 \\
Versae & (5)  & 	\texttt{ENG}	& 42.37	& 49.72	& 45.75	& 1,239	& 1,454	& 616 \\
 &  & 	\texttt{OTHER}	& 35.00	& 15.22	& 21.21	& 46	& 20	& 7 \\
\hline
&  & 	ALL	& 52.17	& 39.22	& 44.78	& 1,285	& 966	& 504 \\
Mgrafu & (6)  & 	\texttt{ENG}	& 52.30	& 40.36	& 45.56	& 1,239	& 956	& 500 \\
 &  & 	\texttt{OTHER}	& 40.00	& 8.69	& 14.29	& 46	& 10	& 4 \\
\hline
&  & 	ALL	& 70.29	& 28.72	& 40.77	& 1,285	& 525	& 369 \\
BERT4EVER & (7)  & 	\texttt{ENG}	& 70.38	& 29.54	& 41.61	& 1,239	& 520	& 366 \\
 &  & 	\texttt{OTHER}	& 60.00	& 6.52	& 11.76	& 46	& 5	& 3 \\
\hline
&  & 	ALL	& 69.92	& 26.23	& 38.14	& 1,285	& 482	& 337 \\ 
BERT4EVER & (8)  & 	\texttt{ENG}	& 70.02	& 26.96	& 38.93	& 1,239	& 477	& 334 \\ 
 &  & 	\texttt{OTHER}	& 60.00	& 6.52	& 11.76	& 46	& 5	& 3 \\
\hline
&  & 	ALL	& 70.49	& 25.84	& 37.81	& 1,285	& 471	& 332 \\
BERT4EVER & (9)  & 	\texttt{ENG}	& 70.60	& 26.55	& 38.59	& 1,239	& 466	& 329 \\
 &  & 	\texttt{OTHER}	& 60.00	& 6.52	& 11.76	& 46	& 5	& 3 \\
\bottomrule
\end{tabular}
\caption{Results on the lower-cased version of the test set.}\label{tab:test-lowercased}
\end{table*}

\subsection{Dataset} 

A corpus of newspaper articles written in Spanish was distributed to the task participants. The corpus articles were sourced from various Spanish newspapers and online media based in Spain. The articles were annotated with unassimilated lexical borrowings.
% TODO: Elena, can we give a list of sources?

Given that lexical borrowings can be multiword expressions (such as \textit{best seller, big data}) and that those units should be treated as one borrowing and not as two independent borrowings, BIO encoding was used to denote the boundaries of each span.

Two classes were used for borrowings: \texttt{ENG} for English borrowings, and \texttt{OTHER} for lexical borrowings from other languages. Tokens that were not part of a borrowing were annotated with the ``outside'' tag (\texttt{O}). Only unassimilated lexical borrowings were considered borrowings. This means that borrowings that have already gone through orthographical adaption (such \textit{fútbol} or \textit{hackear}) were not considered borrowings and were therefore annotated as \texttt{O}. Annotation  guidelines were also made available for participants.

The data was distributed in CoNLL format. An additional collection of documents that was not evaluated (the background set) was released as a part of the test set. This was done to encourage scalability to larger data collections and to ensure that participating teams were not be able to easily perform manual examination of the evaluated part of the test set.

The dataset contained a high number of unique borrowings and OOV words, and there was minimal overlap between splits. This enabled a more rigorous evaluation of system performance, as it helped us better assess the generalizing abilities of the participants' models. Table \ref{tab:corpus}
contains the number of tokens and borrowing spans per type in each split. 

\begin{table*}[t]
\small\centering
\begin{tabular}{lclrrrrrr}
\multicolumn{1}{c}{\textbf{Team}} 
& \multicolumn{1}{c}{\textbf{System}}	
& \multicolumn{1}{c}{\textbf{Type}}
& \multicolumn{1}{c}{\textbf{Prec.}}	
& \multicolumn{1}{c}{\textbf{Rec.}}	
& \multicolumn{1}{c}{\textbf{F1}}
& \multicolumn{1}{c}{\textbf{Ref.}}	
& \multicolumn{1}{c}{\textbf{Pred.}}	
& \multicolumn{1}{c}{\textbf{Corr.}} 
\\
\toprule
 &  & 	ALL	& 90.35	& 82.33	& 86.16	& 1,285	& 1,171	& 1,058\\
Marrouviere &  (1) & 	\texttt{ENG}	& 91.18	& 83.45	& 87.15	& 1,239	& 1,134	& 1,034\\
 &  & 	\texttt{OTHER}	& 64.86	& 52.17	& 57.83	& 46	& 37	& 24\\
\hline
 &  & 	ALL	& 88.71	& 80.08	& 84.17	& 1,285	& 1,160	& 1,029\\
Versae & (2)  & 	\texttt{ENG}	& 90.19	& 81.60	& 85.68	& 1,239	& 1,121	& 1,011\\
 &  & 	\texttt{OTHER}	& 46.15	& 39.13	& 42.35	& 46	& 39	& 18\\
\hline
 &  & 	ALL	& 90.84	& 66.38	& 76.71	& 1,285	& 939	& 853\\
Marrouviere & (3) & 	\texttt{ENG}	& 91.09	& 67.64	& 77.63	& 1,239	& 920	& 838\\
 &  & 	\texttt{OTHER}	& 78.95	& 32.61	& 46.15	& 46	& 19	& 15\\
\hline
 & & 	ALL	& 91.39	& 60.31	& 72.67	& 1,285	& 848	& 775\\
Marrouviere & (4)  & 	\texttt{ENG}	& 92.75	& 61.99	& 74.31	& 1,239	& 828	& 768\\
 &  & 	\texttt{OTHER}	& 35.00	& 15.22	& 21.21	& 46	& 20	& 7\\
\hline
 &  & 	ALL	& 62.76	& 46.30	& 53.29	& 1,285	& 948	& 595\\
Versae & (5) & 	\texttt{ENG}	& 62.97	& 47.62	& 54.23	& 1,239	& 937	& 590\\
 &  & 	\texttt{OTHER}	& 45.45	& 10.87	& 17.54	& 46	& 11	& 5\\
\hline
 & & 	ALL	& 66.81	& 36.50	& 47.21	& 1,285	& 702	& 469\\
Mgrafu & (6)  & 	\texttt{ENG}	& 67.00	& 37.53	& 48.11	& 1,239	& 694	& 465\\
 &  & 	\texttt{OTHER}	& 50.0	& 8.69	& 14.81	& 46	& 8	& 4\\
\hline
 & & 	ALL	& 78.37	& 25.37	& 38.33	& 1,285	& 416	& 326\\
BERT4EVER & (7)  & 	\texttt{\texttt{ENG}}	& 78.40	& 26.07	& 39.13	& 1,239	& 412	& 323\\
 &  & 	\texttt{OTHER}	& 75.00	& 6.52	& 12.00	& 46	& 4	& 3\\
\hline
 & & 	ALL	& 79.03	& 22.88	& 35.49	& 1,285	& 372	& 294\\
BERT4EVER & (8)  & 	\texttt{\texttt{ENG}}	& 79.08	& 23.49	& 36.22	& 1,239	& 368	& 291\\
 &  & 	\texttt{OTHER}	& 75.00	& 6.52	& 12.00	& 46	& 4	& 3\\
\hline
 &  & 	ALL	& 79.34	& 22.41	& 34.95	& 1,285	& 363	& 288\\
BERT4EVER & (9)  & 	\texttt{\texttt{ENG}}	& 79.39	& 23.00	& 35.67	& 1,239	& 359	& 285\\
 &  & 	\texttt{OTHER}	& 75.00	& 6.52	& 12.00	& 46	& 4	& 3\\
\bottomrule
\end{tabular}
\caption{Results on the unquoted version of the test set.}\label{tab:test-unquoted}
\end{table*}

\subsection{Evaluation metrics} 

The evaluation metrics used for the task was the standard precision, recall and F1 over spans:

\begin{itemize}
    \item Precision: The percentage of borrowings in the system’s output that are correctly recognized and classified.
    \item Recall: The percentage of borrowings in the test set that were correctly recognized and classified.
    \item F1-measure: The harmonic mean of Precision and Recall.
\end{itemize}

F1-measure was used as the oﬀicial evaluation score for the final ranking of the participating teams. Evaluation was done exclusively at the span level. This means that only exact matches were considered, and no credit was given to partial matches. For example, given the multitoken borrowing \textit{late night}, the entire phrase would have to be correctly labeled in order to count as a true positive. This makes the evaluation more rigorous, as it avoids the overly-generous scores that can sometimes result from token level evaluation. A model that can only detect English function words would detect \textit{on} and \textit{the} in \textit{on the rocks} or \textit{by} in \textit{stand by} and still get a generous result on a token-level evaluation. 

\subsection{Resource limitation for model training}

The following limitations were established for participants during training:

\begin{itemize}
    \item No additional human annotation  was allowed for training. Given that the main purpose of the shared task was to evaluate how different models perform for the task of borrowing detection, using external data annotated with borrowings would prevent a fair evaluation of different model approaches.
    \item Although the usage of regular lexicons and linguistic resources was accepted, no automatically-compiled lexicons of borrowings (such as those produced by already-existing models that perform borrowing extraction) were allowed. The reason for this limitation was that we were interested in evaluating how different approaches to borrowing detection performed when dealing with previously unseen borrowings, and models that piggyback on already-existing systems's output would prevent that.
\end{itemize}

\begin{table*}[t]
\small\centering
\begin{tabular}{lclrrrrrr}
\multicolumn{1}{c}{\textbf{Team}} 
& \multicolumn{1}{c}{\textbf{System}}	
& \multicolumn{1}{c}{\textbf{Type}}
& \multicolumn{1}{c}{\textbf{Prec.}}	
& \multicolumn{1}{c}{\textbf{Rec.}}	
& \multicolumn{1}{c}{\textbf{F1}}
& \multicolumn{1}{c}{\textbf{Ref.}}	
& \multicolumn{1}{c}{\textbf{Pred.}}	
& \multicolumn{1}{c}{\textbf{Corr.}} 
\\
\toprule
 & & 	ALL & 	78.04	& 82.96	& 80.42	& 1,285	& 1,366	& 1066\\
Marrouviere & (1)  & 	\texttt{\texttt{ENG}}  & 	78.67	& 83.94	& 81.22	& 1,239	& 1,322	& 1040\\
 &  & 	\texttt{OTHER}  &  59.09	& 56.52	& 57.78	& 46	& 44	& 26\\
\hline
 &  & 	ALL & 	77.96	& 67.70	& 72.47	& 1,285	& 1,116	& 870\\
Marrouviere & (3) & 	\texttt{\texttt{ENG}}  & 	78.28	& 68.93	& 73.30	& 1,239	& 1,091	& 854\\
 &  & 	\texttt{OTHER}  & 	64.00	& 34.78	& 45.07	& 46	& 25	& 16\\
\hline
 &  & 	ALL & 	81.14	& 61.95	& 70.26	& 1,285	& 981	& 796\\
Marrouviere & (4) & 	\texttt{\texttt{ENG}}  & 	82.36	& 63.68	& 71.83	& 1,239	& 958	& 789\\
 &  & 	\texttt{OTHER}  &  30.43	& 15.22	& 20.29	& 46	& 23	& 7\\
\hline
 &  & 	ALL & 	60.07	& 81.48	& 69.15	& 1,285	& 1,743	& 1,047\\
Versae & (2) & 	\texttt{\texttt{ENG}}  & 	61.76	& 83.05	& 70.84	& 1,239	& 1,666	& 1,029\\
 &  & 	\texttt{OTHER}  & 	23.38	& 39.13	& 29.27	& 46	& 77	 & 18\\
\hline
 &  & 	ALL & 	42.41	& 48.48	& 45.24	& 1,285	& 1,469	& 623\\
Versae & (5) & 	\texttt{\texttt{ENG}}  & 	42.48	& 49.72	& 45.82	& 1,239	& 1,450	& 616\\
 &  & 	\texttt{OTHER}  & 	36.84	& 15.22	& 21.54	& 46	& 19	& 7\\
\hline
 &  & 	ALL & 	54.56	& 37.74	& 44.62	& 1,285	& 889	& 485\\
Mgrafu & (6)  & 	\texttt{\texttt{ENG}}  & 	54.60	& 38.82	& 45.38	& 1,239	& 881	 & 481\\
 &  & 	\texttt{OTHER}  & 	50.0	& 8.69	& 14.81	& 46	& 8	 & 4\\
\hline
 &  & 	ALL & 	72.96	& 26.46	& 38.83	& 1,285	& 466	& 340 \\
BERT4EVER & (7)  & 	\texttt{\texttt{ENG}}  & 	72.79	& 27.20	& 39.60	& 1,239	& 463	& 337 \\
 &  & 	\texttt{OTHER}  & 	100	& 6.52	& 12.24	& 46	& 3	& 3 \\ 
\hline
 &  & 	ALL & 	72.75	& 23.89	& 35.97	& 1,285	& 422	& 307 \\
BERT4EVER & (8)  & 	\texttt{\texttt{ENG}}  & 	72.55	& 24.54	& 36.67	& 1,239	& 419	& 304 \\
 &  & 	\texttt{OTHER}  & 	100	& 6.52	& 12.24	& 46	& 3	& 3 \\
\hline
 &  & 	ALL & 	73.17	& 23.35	& 35.40	& 1,285	& 410	& 300 \\
BERT4EVER & (9) & 	\texttt{\texttt{ENG}}  & 	72.97	& 23.97	& 36.09	& 1,239	& 407	& 297 \\
 &  & 	\texttt{OTHER}  & 	100	& 6.52	& 12.24	& 46	& 3	& 3 \\
\bottomrule
\end{tabular}
\caption{Results on the unquoted and lower-cased version of the test set.}\label{tab:test-lowercased-unquoted}
\end{table*}

\section{System descriptions}

We received nine submissions from four different teams. However, only two teams submitted system descriptions. As a result, we have no description whatsoever for two of the participating systems, including the one that obtained the best results. We provide a brief summary of the two participating systems for which we received a submission, and refer the reader to their respective task description papers for further details.

\subsection{BERT4EVER team: CRF model with data augmentation}

The BERT4EVER team \cite{jiang2021bert4ever} submitted a system to ADoBo based on combining several CRF models trained on different portions of the task's training data. The models were used to label a freely-available open corpus in Spanish, and individual models were then re-trained on the output. Results suggest that this strategy improves two F1 points on the test set when compared to a trained-on-task-data-only baseline. The paper combines two well-known items in the ML toolbox, namely CRFs and data augmentation, and shows that bootstrapping an additional dataset is indeed useful. 

\subsection{Versae team: using STILTs}

The Versae team \cite{de2021futility} submitted a system that experimented with using STILTs---supplementary training on intermediate label-data tasks \cite{phang2019sentence}---for the ADoBo task.
They experimented with training using part of speech, named entity recognition, code-switching, and language identification datasets, but found that models trained in this way consistently perform worse than fine-tuning multilingual language models.
The Versae team also explored which multilingual language models perform best, evaluating multilingual BERT, RoBERTa, and models trained on small sets of languages.

\section{Results}

Results of the task were computed using \texttt{SeqScore}\footnote{\url{https://github.com/bltlab/seqscore}} \cite{seqscore}, a Python package for evaluating sequence labeling tasks, configured to emulate the \texttt{conlleval} evaluation script. Scores are summarized in Table~\ref{tab:test}. 
F1 ranged from 37.29 to 85.03, with the Marrouviere team scoring highest (F1=85.03, P=88.81 and R=81.56), close to the next-highest scores from the Versae team (F1=84.80, P=88.77 and R=81.17). 

In order to get a better understanding of the systems that took part in the shared task, we performed some experiments on the output that was submitted by participants.

\subsection{Combining outputs}

In order to assess the complementarity of the submitted systems, an experiment was carried out combining their outputs. The combination consisted of the union of all detected terms. Since the number of systems is not very high, all combinations of systems were explored. In terms of F1 score, the best performing combination was (1), (2), and (4), with F1=87.83, P=87.83, and R=89.26, a result that outperforms the scores obtained separately by each individual system.

\subsection{Removing ortho-typographic cues}

Three variations of the test set were included in the background set (the additional collection of documents released along with the test set): 

\begin{enumerate}
    \item A lowercase version, where all uppercase letters in the original test set were transformed to lowercase.
    \item A no-quotation-mark version, where all quotation marks in the original test set (`` '' ` ' « ») were removed.
    \item A lowercase no-quotation-mark version, where all uppercase letters where transformed to lowercase AND all quotation marks were removed.
\end{enumerate}

None of these versions were used to rank the systems but to observe the systems difference in performance on different textual characteristics. The rationale for these experiments was to assess how well systems  performed if certain orthotypographic cues that usually appear along with borrowings (such as quotation marks) were removed. After all, a borrowing is still a borrowing regardless of whether it is written with or without quotation marks and it would be of little use to have a model that systematically labeled anything between quotation marks as a borrowing, or that only detected borrowings if they are written between quotation marks.

Similarly, many of the foreign words that appear in newswire are usually proper names, where the uppercase can serve as cue to distinguish them from borrowings. Given that speakers are capable of distinguishing borrowings from proper names in oral settings---where no case distinction exists---and that these cues are not present in other textual genres (e.g. social media), we were interested in assessing how well the models performed when no case cue was available.

Results for these experiments are presented in Tables~\ref{tab:test-lowercased}, \ref{tab:test-unquoted} and \ref{tab:test-lowercased-unquoted}. Focusing on the best two performing systems, we observe a drop of global F1 due to a consistent drop on precision not compensated with the a slightly increase of recall for the lowercased versions of the test set. In general, the drop in system (2) is more pronounced than in system (1), which causes its repositioning in the corresponding rankings. For the unquoted version of the test set, system (1) increases its F1 and system (2) decrements it slightly. Not having information on system (1), we can not attribute any of the differences to any characteristics of the systems.

\section{Conclusions}

In this paper we have presented the results of the ADoBo shared task on extracting unassimilated lexical borrowings from Spanish newswire. We have introduced the motivation for this topic, we have described the scope and nature of the proposed task, we have shared the obtained results and have summarized the main findings. Participants results ranged from F1 scores of 37 to 85. These scores show that this is not a trivial task and that lexical borrowing detection is an open problem that requires further research.

Our goal with this shared task was to raise awareness about a topic that, although highly relevant in the linguistics literature, has been mostly neglected within NLP. Although the participation for this first edition was modest (nine systems submitted from four different teams), the response was positive and it seems to indicate that there exists a moderate population within the community that is interested in borrowing as an NLP task. In fact, a post-task survey distributed among registered participants showed that 85\% of respondents were interested in seeing future editions around this phenomenon, particularly on languages other than Spanish and including both semantic and diachronic borrowings.

\bibliographystyle{fullname}
\bibliography{EjemploARTsepln}

\end{document}